\documentclass[sigconf]{acmart} 

\usepackage{fancyhdr}
\usepackage{graphicx}
\usepackage{multirow}
\usepackage{bbding}
\usepackage{float}
\renewcommand\footnotetextcopyrightpermission[1]{}
\AtBeginDocument{%
  \providecommand\BibTeX{{%
    \normalfont B\kern-0.5em{\scshape i\kern-0.25em b}\kern-0.8em\TeX}}}

\begin{document}

\title{Understanding Mobile GUI: from Pixel-Words to Screen-Sentences}


\author{Jingwen Fu}
\authornote{Equal contributions during internship at Microsoft Research Asia.}
\affiliation{%
  \institution{Xi'an Jiaotong University}
  \city{Xi'an}
  \country{China}}
\email{fu1371252069@stu.xjtu.edu.cn}

\author{Xiaoyi Zhang}
\authornotemark[1]
\affiliation{%
  \institution{Microsoft Research Asia}
  \city{Beijing}
  \country{China}
}
\email{theyaoyi626@gmail.com}

\author{Yuwang Wang}
\authornote{Corresponding author}
\affiliation{%
 \institution{Microsoft Research Asia}
 \city{Beijing}
 \country{China}}
\email{yuwang.wang@microsoft.com}

\author{Wenjun Zeng}
\affiliation{%
  \institution{Microsoft Research Asia}
  \city{Beijing}
  \country{China}}
 \email{wezeng@microsoft.com}
 
 \author{Sam Yang}
\affiliation{%
  \institution{Microsoft}
  \city{Redmond}
  \country{USA}}
 \email{samyang@microsoft.com}
 
 \author{Grayson Hilliard}
\affiliation{%
  \institution{Microsoft}
  \city{Redmond}
  \country{USA}}
 \email{wihillia@microsoft.com}

\renewcommand{\shortauthors}{Jingwen Fu, Xiaoyi Zhang, Yuwang Wang, et al.}


\begin{abstract}
The ubiquity of mobile phones makes mobile GUI understanding an important task. Most previous works in this domain require human-created metadata of screens (e.g. View Hierarchy) during inference, which unfortunately is {often not available or reliable enough for GUI understanding}. Inspired by the impressive success of Transformers in NLP tasks, targeting for purely vision-based GUI understanding, we extend the concepts of \emph{Words}/\emph{Sentence} to \emph{Pixel-Words}/\emph{Screen-Sentence}, and propose a mobile GUI understanding architecture: \emph{Pixel-Words} to \emph{Screen-Sentence} (PW2SS). In analogy to the individual \emph{Words}, we define the \emph{Pixel-Words} as \textbf{atomic} visual components (text and graphic components), which are visually consistent and semantically clear across screenshots of a large variety of design styles.
The \emph{Pixel-Words} extracted from a screenshot are aggregated into \emph{Screen-Sentence} with a Screen Transformer proposed to model their relations. 
Since the \emph{Pixel-Words} are defined as atomic visual components, the ambiguity between their visual appearance and semantics is dramatically reduced. We are able to make use of metadata available in training data to auto-generate high-quality annotations for \emph{Pixel-Words}. A dataset, RICO-PW, of screenshots with \emph{Pixel-Words} annotations is built based on the public RICO dataset, which will be released to help to address the lack of high-quality training data in this area.
We train a detector to extract \emph{Pixel-Words} from screenshots on this dataset and achieve metadata-free GUI understanding during inference.
We conduct experiments and show that \emph{Pixel-Words} can be well extracted on RICO-PW and well generalized to a new dataset, P2S-UI, collected by ourselves. The effectiveness of  PW2SS is further verified in the GUI understanding tasks including relation prediction, clickability prediction, screen retrieval, and app type classification.
\end{abstract}




\maketitle

\begin{figure}[t]
\centering
\includegraphics[width=0.48\textwidth]{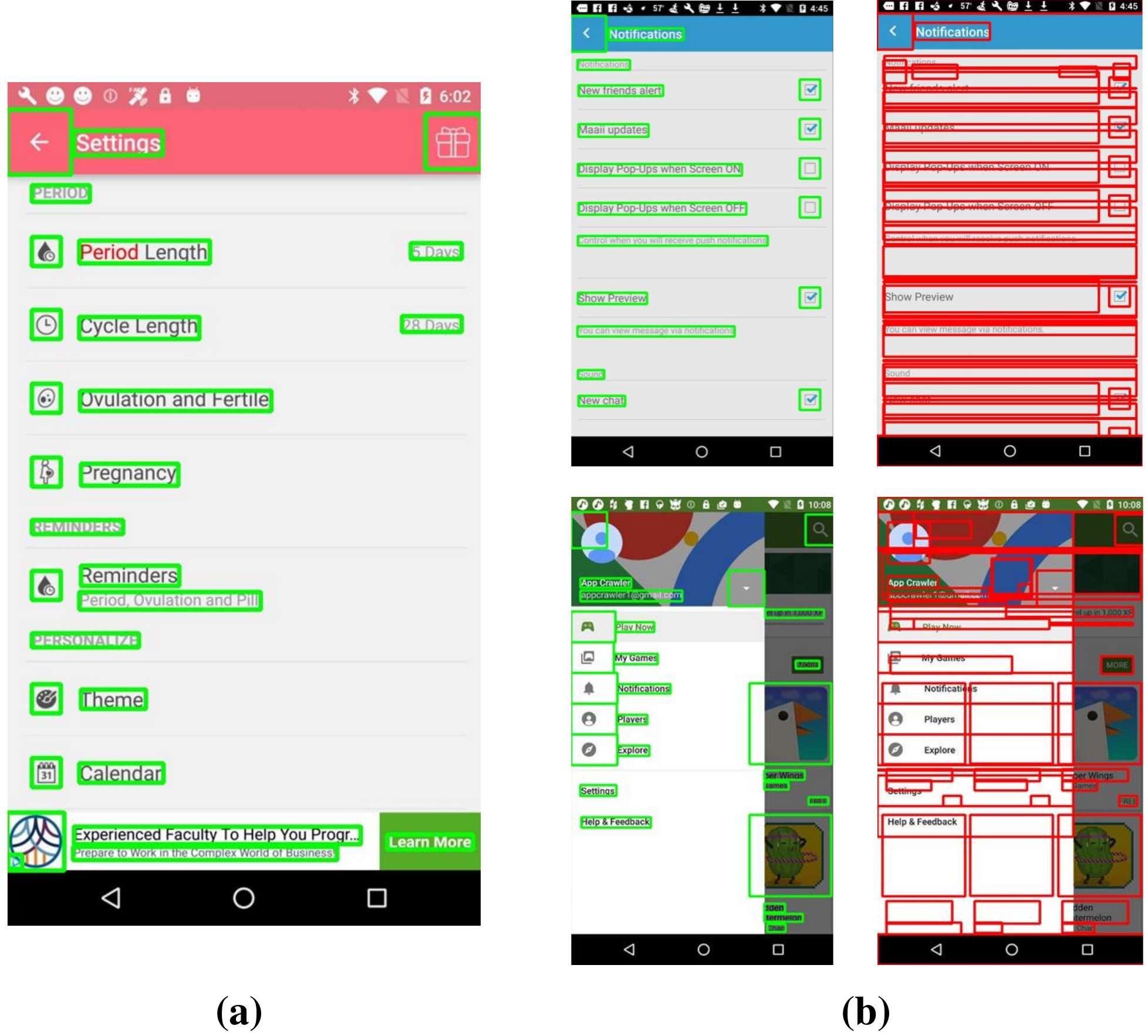} 
\caption{
(a) Visualization of our proposed \emph{Pixel-Words} (labeled with green boxes). (b) Compared to leaf nodes in VH (labeled with red boxes), \emph{Pixel-Words} are more clean and visually consistent across screenshots.
}
\label{fig:pixel-words}
\vspace{-5mm}
\end{figure}

\section{Introduction}

As mobile phones have become indispensable for human daily life, understanding GUI becomes a very important capability for AI to accomplish such tasks  as language navigation~\cite{li2020mapping}, task automation~\cite{li2021screen2vec,zhang2021screen}, reverse software engineering~\cite{beltramelli2018pix2code,nguyen2015reverse}, screen reader~\cite{chen2020unblind}, etc. 
The metadata of screenshot, e.g. View Hierarchy (VH), provides a tree structured description of the UI components forming the screenshot. Most previous works \cite{he2020actionbert}\cite{li2020mapping}\cite{li2021screen2vec} rely on the metadata to understand GUI. Unfortunately, the metadata is often noisy\cite{zhang2017interaction,zhang2018robust,zhang2021screen} due to the large varieties of platforms, third-party UI tookits, coding styles, etc~\cite{zhang2021screen}. What's worse is that the metadata is often not accessible due to privacy or compliance issues. Understanding the GUI from the screenshot only is a challenging and less-studied area. The first challenge is the complexity of screenshots. Different from natural images, the screenshots consist of UI components of a large number of categories. The UI components from different categories can be visually similar, and complex UI components may be composed of other simpler UI components, e.g., list views can be decomposed into icons and texts. The visual appearances of UI components vary due to the large variety of UI design styles. Chen et al.\cite{chen2020object} try to detect the non-text UI components from screenshots with traditional low-level vision algorithms, e.g. boundary extraction, but they lack semantic understanding and are not robust on screenshots with complex layouts. Zhang et al.~\cite{zhang2021screen} manually annotate the UI components of a self-collected private dataset and train their detector to detect those UI components. It is difficult to extend the human labelling method to larger scale data and there is a mismatch between human annotations and metadata for various kinds of UI components. Besides, both works are limited to only detecting UI components from screenshots, without screen-level understanding.
Another challenge is the lack of high-quality annotated large scale datasets, which limits deep learning based methods. Liu et al.~\cite{liu2018learning} aim to generate the annotations of UI components by parsing the metadata with hand-crafted rules, but the quality is limited by the noisy metadata. 

Our work is inspired by the modeling ideas in NLP. Each individual \emph{Word} is an atomic component with the essential semantics, which is modeled as a token. All the tokens in a \emph{Sentence} are fed into a Transformer to understand the whole \emph{Sentence}. The key is to achieve the understanding of the whole sequence from the basic units. This successful modeling framework can be extended to understand UI. In analogy to NLP, the ``word" of a screen should be: 1) atomic visual components carrying clear semantic meaning. 2) visually consistent across different UIs. A VH based work~\cite{he2020actionbert} takes the leaf nodes in metadata as the words for screen. However, as shown in Figure~\ref{fig:pixel-words} (b), those leaf nodes are very noisy and some nodes contains only icons or texts, while some other leaf nodes contain both. The confusing semantics of the leaf nodes would hurt the understanding of individual components and the whole UI.  

In this paper, we aim to achieve GUI understanding from pixels by extending the \emph{Word}/\emph{Sentence} concepts into the \emph{Pixel-Word}/\emph{Screen-Sentence} concepts, and propose a new architecture: from \emph{Pixel-Words} to \emph{Screen-Sentence} (PW2SS). Our PW2SS gets rid of the limitation of requiring metadata in inference and can be widely applied across different platforms, UI design tools and styles. The key is to design the \emph{Pixel-Words} to have the property of \emph{Words} in NLP. 
We define the \emph{Pixel-Words} as atomic visual components of screens, which include \emph{Text Pixel-Words} (text) extracted by OCR and \emph{Graphic Pixel-Words} (icons and images) extracted by our Graphic Detector as shown in Figure~\ref{fig:pixel-words}. The appearances of those \emph{Pixel-Words} are clean and consistent across different screenshots. The benefits of our \emph{Pixel-Words} definition are three-fold: 1) providing semantically effective tokens into the Transformer to understand the Screen-Sentences. 2) Enabling the OCR and Graphic Detector to extract \emph{Pixel-Words} based on the consistent visual appearance of these components. 3) Making it possible to clean the noise in the metadata and get pseudo-labels of the \emph{Pixel-Words} for the training of the detector. According to the above definition of \emph{Pixel-Words}, we propose a heuristic method to generate pseudo label for \emph{Pixel-Words} and build a \emph{Pixel-Words} dataset named RICO-PW based on the pulished RICO dataset \cite{deka2017rico}. 
For \emph{Screen-Sentence}, we leverage BERT\cite{devlin2018bert} to design a Screen Transformer to model the relation of \emph{Pixel-Words} and train the Screen Transformer with masked \emph{Pixel-Words}
prediction. 

To evaluate the effectiveness of our \emph{Pixel-Words} and PW2SS, we conduct experiments on RICO-PW and a new GUI understanding dataset, P2S-UI, collected by ourselves. Our method achieves the best performance on the GUI understanding tasks including \emph{Pixel-Words} extraction, relation prediction, clickability prediction, screen retrieval, and app type classification.


Our main contributions can be summarized as follows:
\begin{itemize}
    \item We propose a pixel-based GUI understanding framework from pixels which is suitable for general applications across different platforms, UI design tools and styles.
    \item By defining the \emph{Pixel-Words} as atomic text and graphic components, we make them visually consistent and semantically clear tokens for the Screen Transformer. 
    \item We build a high-quality UI datasets RICO-PW with \emph{Pixel-Words} annotations, which will be released to the public.
\end{itemize}

\begin{figure*}[t]
\centering
\includegraphics[width=0.8\textwidth]{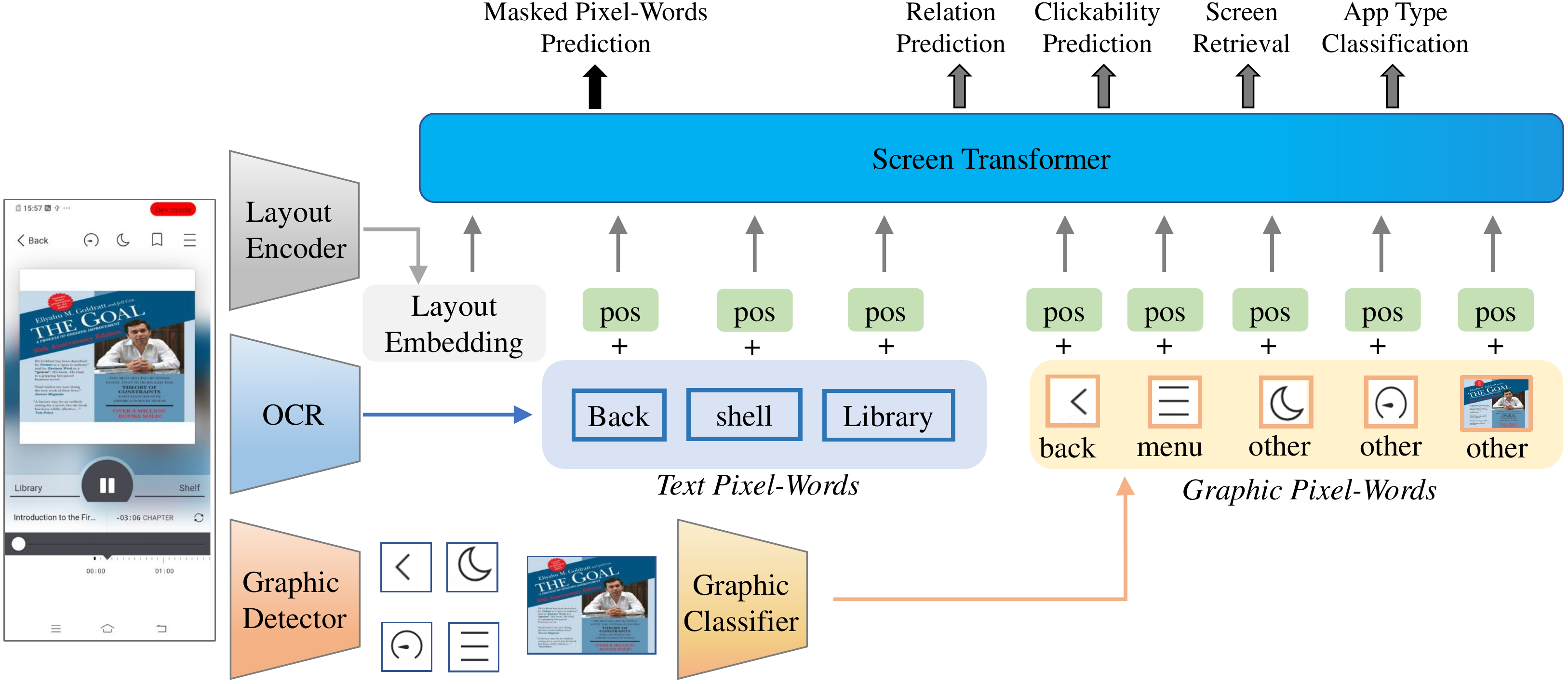} 
\caption{Overview of our method. Given a screenshot, we first generate \emph{Pixel-Words} with an OCR module to extract \emph{Text Pixel-Words} and an Graphic Detector and Graphic Classifier to extract \emph{Graphic Pixel-Words}. We also use a Layout Encoder to get a Layout Embedding proving a global layout information. All the \emph{Pixel-Words} and the layout embedding are fed into the Screen Transformer to model the \emph{Screen-Sentence}.}
\label{fig:overview}
\end{figure*}

\section{Related Work}
\textbf{GUI Understanding} is a challenging area which has attracted increasing attention recently due to the prevalence and importance of smart devices. One important task is extracting elements from screenshots and learning the representations of the components and the whole screen. The related works can be divided into two branches: purely vision-based methods which only take the screenshot as input and metadata-based methods which require the metadata of the screenshot as an extra input. The first branch ~\cite{chen2020object,zhang2021screen,sun2020ui} tries to extract UI components from screenshots with detection techniques, but they are limited to component level understanding, and lack the understanding of the relation between components and the whole screen. Besides, their detection ground truth are extracted from metadata directly, and the visual ambiguity of different UI components may hurt the detection performance. The second branch achieves screen level understanding with the help of metadata. Seq2act \cite{li2020mapping} retrieves elements in screen using structured query. ActionBert \cite{he2020actionbert} and Screen2Vec \cite{li2021screen2vec} learn the embeddings of elements and screens. However, {these methods depend on the metadata, which are not universally accessible and often suffer from noise.}

\noindent
\textbf{Transformer-based representation learning} aims to model the relations of tokens with Transformers and provide a global understanding of the whole input. The Transformer is proposed in \cite{vaswani2017attention} and BERT\cite{devlin2018bert} is a well-accepted pretraining method for transformer. Recently, many works apply Transformers to visual language tasks \cite{huang2020pixel,lu2019vilbert,li2020unicoder,su2019vl}. Our work is different from these works in that they require extra text information, e.g. image caption, for each image, but our text information is extracted from screenshots by the OCR. The most related work to ours is LayoutLM \cite{xu2020layoutlm,xu2020layoutlmv2} for form understanding. The main difference is that the graphic components like icons and images are more important in UI understanding, while LayoutLM mainly considers the text in the scanned document. 

\noindent
\textbf{Mobile screen related applications} focus on applying computer vision techniques in mobile UIs to solve particular tasks. Due to the ubiquity of the mobile phone, there are a lot of applications being explored. Reverse software engineering \cite{chen2018ui,beltramelli2018pix2code,nguyen2015reverse,moran2018machine} aims to generate the code of UI from the corresponding screenshot. Design assistant~\cite{bunian2021vins,chen2020wireframe,zhao2021guigan} help to design a GUI of apps. Assistants for the visually impaired \cite{chen2020unblind} aims to generate description of the element. The previous works \cite{liu2020owl,cooper2021takes} detect the displaying issue in mobile screens with computer vision technique. These papers mainly focus on the specific application scenario. 
Those tasks can be better accomplished with a good understanding of the screens and can be regarded as the downstream tasks of our work. 

\noindent
\textbf{Mobile screen related datasets} are very important for the training of models of GUI understanding. For deep learning based methods, large scale datasets are necessary for reliable performance and generalizability. RICO~\cite{deka2017rico} is one of the most important public datasets with both screenshots dataset with corresponding metadata in this area. However, it only contains some simple auto-generated labels without semantics. Liu et al. \cite{liu2018learning} propose a way to generate the semantic labels from the metadata, but it is still noisy and is not suitable for the purpose of GUI understanding from pixels. In this paper, we generate high quality annotations of \emph{Pixel-Words} for GUI understanding tasks. 
\section{From Pixel-Words to Screen-Sentences}

\subsection{Overview of PW2SS}
To understand the GUI from pixels, we extent the concepts of \emph{word}/\emph{Sentence} in NLP to \emph{Pixel-Words}/\emph{Screen-Sentences}. 
Different from NLP, where the \emph{Word} is already isolated, for GUI understanding, we need to first extract \emph{Pixel-Words} from the given screenshot, then feed the \emph{Pixel-Words} into our Screen Transformer as shown in Figure~\ref{fig:overview}.  We first train the models to extract high-quality \emph{Pixel-Words} and then train the Screen Transformer. 

In analogy to \emph{Words}, the \emph{Pixel-Words} should be isolated visual components and carry elementary semantics. Screenshot are typically rendered from metadata, e.g. View Hierarchy (VH), describing the hierarchical structure of UI components in the screenshot. (Examples of VH data are shown in our Supplementary Material). One straightforward way to isolate ``words'' of a screen is to use the leaf nodes in VH as \emph{Pixel Words}. However, as shown in Figure~\ref{fig:pixel-words}, due to varieties of UI tools and coding styles, there are many invalid leaf nodes and the content in the leaf node varies across different screens. We define the \emph{Pixel-Words} as the visual atomic components, which make up the screen. As Figure~\ref{fig:pixel-words} (a) shows, our \emph{Pixel-Words} include text components (e.g. text or blocks of texts) and graphic components (e.g. icons and images), denoted as \emph{Text Pixel-Words} and \emph{Graphic Pixel-Words} respectively. Those \emph{Pixel-Words} are both visually consistent and semantically carrying meaningful information. 

Based on the \emph{Pixel-Words}, the Screen Transformer is able to understand the whole screen. We refer to BERT~\cite{devlin2018bert} to add position embeddings to represent the locations of the \emph{Pixel-Words} and use the masked \emph{Pixel-Words} prediction task to pretrain the Screen Transformer. Furthermore, we design a layout embedding to provide global  layout information of screen. Then the pretrained Screen Transformer can be finetuned to accomplish various downstream tasks like relation prediction, clickability prediction, screen retrieval and app type classification etc.


\subsection{\emph{Pixel-Words} Extraction and Understanding}
\label{sec:PW}
We extract the \emph{Pixel-Words} from a screenshot and transfer them into tokens based on different appearances of \emph{Text Pixel-Words} and \emph{Graphic Pixel-Words} as shown in Figure~\ref{fig:overview}. For the \emph{Text Pixel-Words},  we use an off-the-shelf OCR tool\footnote{https://developers.google.com/ml-kit/vision/text-recognition} to extract the location and text content, then feed the text into Sentence-BERT\cite{reimers2019sentence} to get the representation of tokens. For the \emph{Graphic Pixel-Words}, we design a Graphic Detector and Graphic Classifier to extract the location and semantics. The Graphic Detector detects the graphic elements, then we use a Classifier to identify the semantic meaning of the elements, and use the BERT embeddings of semantic labels as the tokens.

\begin{table}[t]
\centering
\resizebox{.95\columnwidth}{!}{
\begin{tabular}{c|c|c|c|c|c} 
\hline
\multirow{2}{*}{Methods} & \multirow{2}{*}{Cleaning} & \multicolumn{2}{c|}{\emph{Graphic Pixel-Word}} & \multicolumn{2}{c}{\emph{Text Pixel-Word}}  \\ 
\cline{3-6}
                        &                           & Recall & Precision                       & Recall & Precision                   \\ 
\hline
Liu et al.              & w/o                       & 0.73   & 0.71                            & 0.69   & 0.62                        \\
Liu et al.              & w/                        & 0.72   & 0.75                            & 0.69   & 0.69                        \\
Ours                    & w/o                       & 0.80   & 0.94                            & 0.86   & 0.97                        \\
Ours                    & w/                        & 0.85   & 0.95                            & 0.95   & 0.97                        \\
\hline
\end{tabular}}
\caption{Comparison the labels generated from the metadata using our method and Liu et al.~\cite{liu2018learning} on 200 screenshots labelled by us. The "cleaning" means the operation of removing the screenshots with metadata.}
\label{tab:refined-label}
\vspace{-5mm}
\end{table}

\begin{figure*}[t]
\centering
\includegraphics[width=0.65\textwidth]{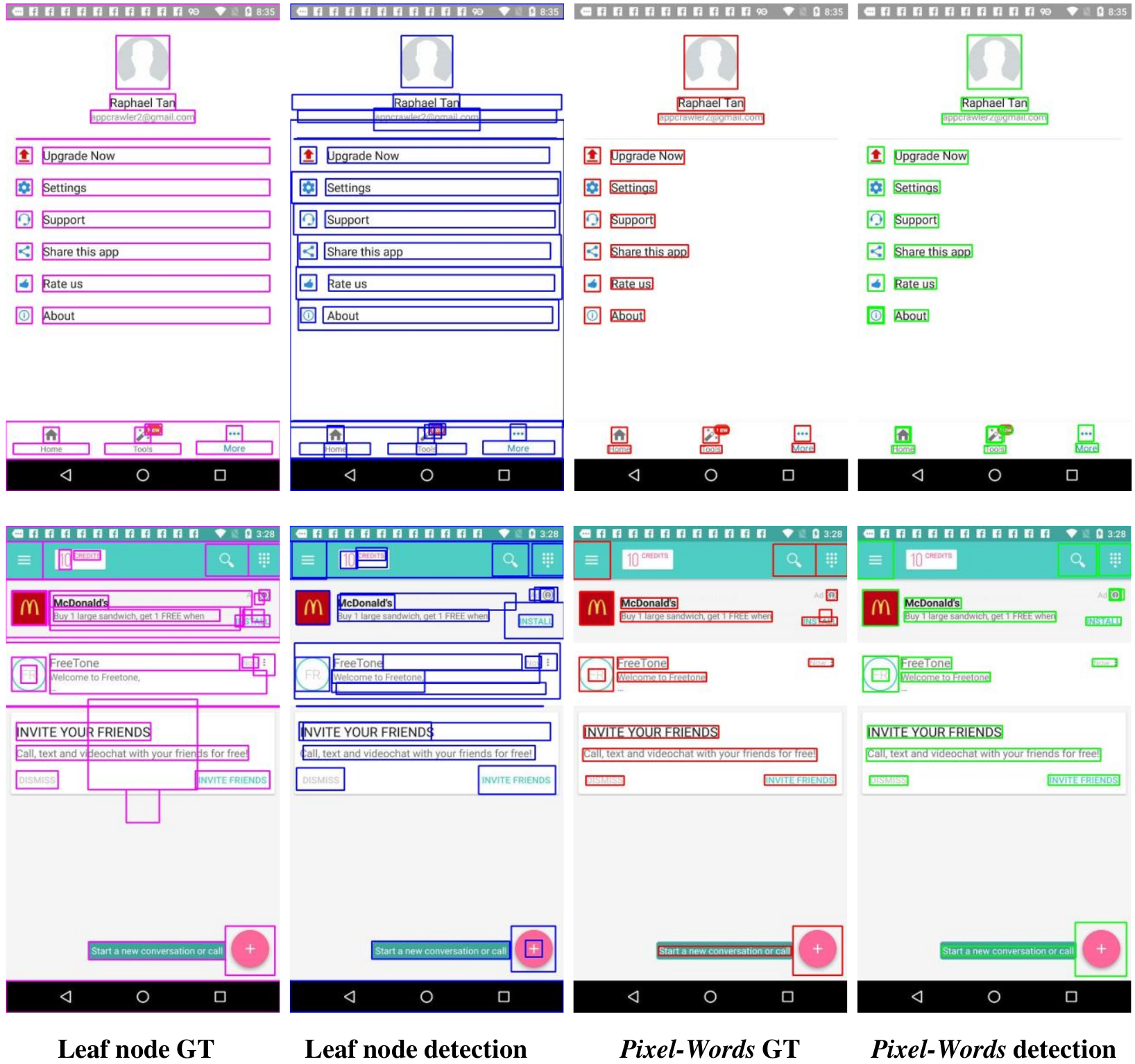} 
\caption{Visualization of leaf nodes GT (magenta), leaf nodes detection results (blue), \emph{Pixel-Words} ground-truth (red) and \emph{Pixel-Words} detection results (green).}
\vspace{-3mm}
\label{fig:vis_det}
\end{figure*}


To train the Graphic Detector with less human labeling cost, we make use of VH in the RICO dataset\cite{deka2017rico, liu2018learning}. The biggest challenge for getting supervision for UI components is that it is hard to filter out invalid nodes using visual features. However, our clear definition of \emph{Graphic Pixel-Words} shows great advantages here. For \emph{Graphic Pixel-Words}, we only focus on the atomic graphic components in the UI which have consistent appearance and can be easily distinguished from other complex UI components. The details are shown in the Supplementary Material. 



Aaccording to our definition of \emph{Pixel-Words}, a baseline method is to leverage the component annotations provided by Liu et al.~\cite{liu2018learning} on RICO. We manually choose the components labelled with text-related categories as our \emph{Text Pixel-Words}, and choose the components labelled with icon- or image-related categories as our \emph{Graphic Pixel-Words}. However, the \emph{Text Pixel-Words} collected in this way are still very noisy due to the low quality annotations of Liu et al.~\cite{liu2018learning}.
To obtain a large number of \emph{Pixel-Words} supervision with low cost of human effort, we carefully design a pipeline to extract annotation from VH. For text, we first extract nodes in VH with UI class names related to "text", then we use the OCR tool to localize text inside these nodes. For graphics, we generate candidates according to location of texts and nodes in VH, then we use a binary classifier to identify each candidate is a graphic or not. We collect 1239 candidate patches from screenshots and manually label them as positive or negative samples. The binary classifier is trained on this dataset and demonstrates high accuracy performance. We further propose a ``cleaning'' operation to filter out invalid screenshots where there is an obvious mismatch between the number of text boxes extracted using OCR or metadata. 

To verify the effectiveness of the generated labels using our method, we manually label the \emph{Pixel-Words} in 200 screenshots randomly sampled from RICO and use them as ground truth. 
As Table~\ref{tab:refined-label} shows, our generated labels for \emph{Pixel-Words} achieve higher Recall and Precision than the leaf nodes of VH.

For further understanding the extracted graphics, we use a Graphic Classifier to recognize the semantic meaning of each graphics, as shown in Figure~\ref{fig:overview}. To train the Graphic Classifier, we build a high quality graphic dataset, named RICO-ICON, by refining category annotations from RICO. Based on the observation that there exists a serious long-tail problem of icon categories in the original RICO dataset, we only collect the most important categories in RICO-ICON. We use the clicking frequency as a metric to measure the importance of graphic categories and select 31 categories with the highest clicking frequency as the graphic categories in RICO-ICON. The remaining categories of graphic are assigned an "other" category label.

\begin{table}[t]
\centering
\resizebox{.7\columnwidth}{!}{
\begin{tabular}{c|c|c|c|l} 
\hline
\multirow{2}{*}{Method} & \multicolumn{4}{c}{Metric}                                  \\ 
\cline{2-5}
                        & AR & AP   & \multicolumn{1}{l|}{AP50} & AP75   \\ 
\hline
Leaf Nodes              & 0.665          & 0.575 & 0.720                     & 0.611  \\
\emph{Pixel-Words}                    & \textbf{0.780}          & \textbf{0.636} & \textbf{0.835}                     & \textbf{0.746}  \\
\hline
\end{tabular}}
\caption{Comparison of the detection results between leaf node and \emph{Pixel-Words} on RICO-PW.\label{tab:leafnode_det}}
\vspace{-5mm}
\end{table}

\subsection{\emph{Screen-Sentence} Understanding}
{To accomplish tasks requiring semantic understanding of the whole screen, we take the \emph{Pixel-Words} from the previous stage as tokens and feed them into the Screen Transformer to model their relation and form a \emph{Screen-Sentence}.}
We design our Screen Transformer as a 6-layer transformer architecture referring to BERT~\cite{devlin2018bert}. As Figure~\ref{fig:overview} shows,  the input of the transformer consists of three parts: \emph{Pixel-Words} embeddings, corresponding 2-D position embeddings and a layout embedding.
\emph{Pixel-Words} embeddings represent the sematics of these atomic visual components. The embeddings of \emph{Text Pixel-Words} and \emph{Graphic Pixel-Words}  are processed by a linear layer to ensure they have the same size.
Position embeddings are added to each corresponding \emph{Pixel-Words} to provide spatial position. Following the same setting as LayoutLM\cite{xu2020layoutlm}, we represent the position and size of \emph{Pixel-Words} in screenshot as $(x_{min},y_{min},x_{max},y_{max},w,h)$, where $w$ and $h$ denote the width and height of the \emph{Pixel-Words} respectively.
A layout embedding is designed to provide the global layout information of the components.  We follow \cite{liu2018learning} to generate layout representation using the bounding boxes of our \emph{Pixel-Words}. 


\begin{table}[t]
\centering
\resizebox{.98\columnwidth}{!}{
\begin{tabular}{c|cccc|cccc} 
\hline
\multirow{2}{*}{models} & \multicolumn{4}{c|}{Val}      & \multicolumn{4}{c}{Test}       \\ 
\cline{2-9}
                        & AR    & AP   & AP50  & AP75  & AR    & AP   & AP50  & AP75   \\ 
\hline
RTN-S                   & 0.720 & 0.613 & 0.753 & 0.658 & 0.720 & 0.609 & 0.748 & 0.654  \\
ATSS-S                  & 0.716 & 0.587 & 0.782 & 0.629 & 0.718 & 0.587 & 0.778 & 0.628  \\
FA-S                    & 0.756 & 0.656 & 0.756 & 0.705 & 0.750 & 0.645 & 0.749 & 0.693  \\ 
\hline
FA-D                    & 0.790 & 0.678 & 0.841 & 0.725 & 0.792 & 0.676 & 0.837 & 0.722  \\
\hline
\end{tabular}}
\caption{Comparison of different detectors' performance on RICO-PW validation and test splits. AR is short for Average Recall. RTN-S, ATSS-S, FA-S  denotes RetinaNet, ATSS, FreeAnchor with ResNet50\cite{he2016deep} as the backbone respectively. FA-D denotes FreeAnchor with ResNeXt101\cite{xie2017aggregated} as backbone.}
\label{tb:det_baselines}
\vspace{-7mm}
\end{table}



The Screen Transformer is pretrained with a self-supervised task: masked \emph{Pixel-Words} prediction. We select four downstream tasks covering the understanding from \emph{Pixel-Words} level to \emph{Sentence-Screen} level. Those tasks can be well used to evaluate the understanding performance and are practically important.

\vspace{1mm}
\noindent 
\textbf{Pretraining Task: Mask \emph{Pixel-Words} prediction} \;
Inspired by BERT\cite{devlin2018bert}, where the input tokens are randomly masked and predicted from the remaining tokens, we design the masked \emph{Pixel-Words} prediction task to regress the embedding of the masked \emph{Pixel-Words}. The purpose is to force the transformer to learn the dependencies among \emph{Pixel-Words} in the screen. More specifically, we randomly mask $15\%$ of the total \emph{Pixel-Words} fed into the Screen Transformer, and take the objective fucntion as $\ell_2$ norm between the masked \emph{Pixel-Words} and predicted \emph{Pixel-Words}.



\vspace{1mm}
\noindent 
\textbf{Downstream task \#1: Clickability Prediction}\;
The task is to predict whether the \emph{Pixel-Word} is clickable or not.  A \emph{Pixel-Word} is clickable when it is a button or part of a button. We use a classifier (a 3-layer MLP) to determine the clickability of the \emph{Pixel-Word} from the embedding output by the Screen Transformer. 

\begin{figure*}[t]
\centering
\includegraphics[width=0.98\textwidth]{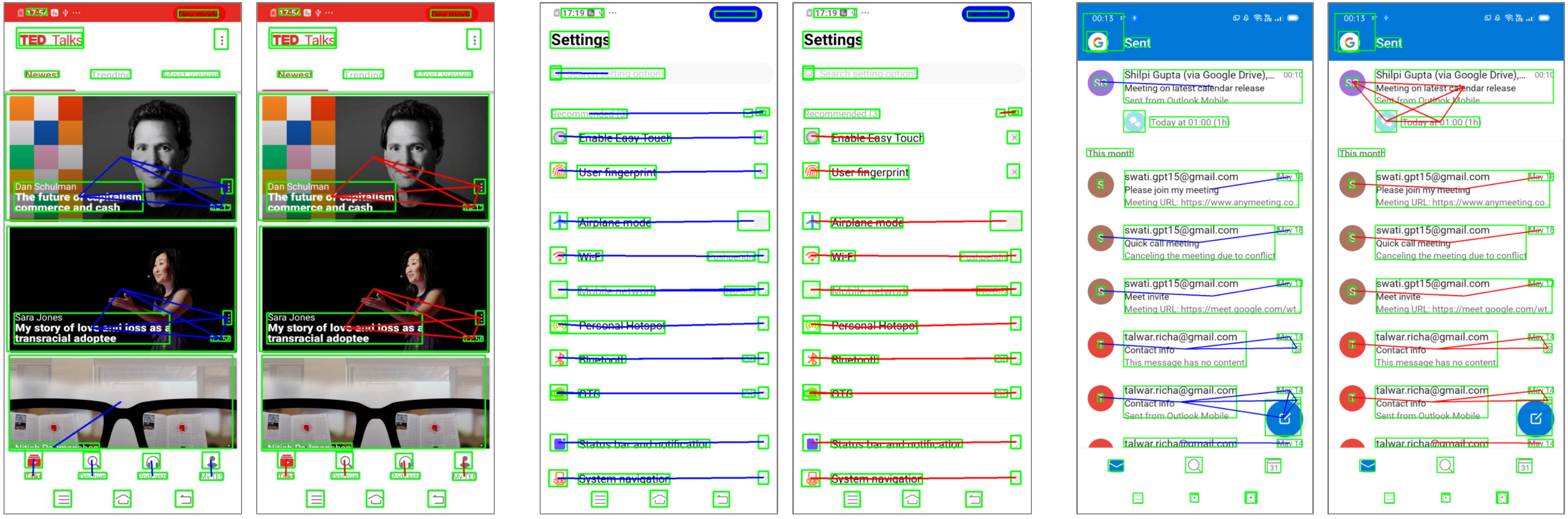} 
\caption{Visualization of relations prediction results of PW2SS on P2S-UI. A line is shown to connect two \emph{Pixel-Words} with relation. The predicted results is labeled with blur lines and ground truth is labeled with red lines. }
\label{fig:RP}
\vspace{-3mm}
\end{figure*}

\vspace{1mm}
\noindent 
\textbf{Downstream task \#2: Relation Prediction}\;
The relation prediction task is to predict the pairwise relation between \emph{Pixel-Word} pairs. The relations reflect how semantically related of the two \emph{Pixel-Words}. For example, as shown in Figure~\ref{fig:RP} (b), the WiFi icon and the texts "WiFi" nearby have the same semantic meaning and they should have close relation. 
We first sum the embeddings of two \emph{Pixel-Words} output from the Screen Transformer and feed the result into a 3-layer multi-layer perceptron (MLP) to predict the relation of the two  \emph{Pixel-Words}.   
The objective function is the cross entropy between the predicted relation and the ground truth.  

\vspace{1mm}
\noindent 
\textbf{Downstream task \#3: Screen Retrieval}\;
In analogy to sentence retrieval in NLP, screen retrieval task aims to find the screenshots that are the most similar to the query screenshot in terms of both the semantics of the content and the layout of the structure. 

To accomplish screen retrieval, a good understanding of the whole screen is required. To obtain a representation for a given screenshot, we apply max pooing on all the outputs of the Screen Transformer. Then we find the closest screenshots for the query screenshot using the cosine similarity of the representations of the screenshots.

\vspace{1mm}
\noindent 
\textbf{Downstream task \#4: App Classification}\;
The app type classification task is to recognize the app type for each screenshot. We also use a maxpooling operation to aggregate all the output embedding from Screen Transformer to get the representation of the given screen. Then we feed the representation into a classifier (a 3-layer MLP) to predict the app type of the screen. 

\begin{table}[t]
\centering
\resizebox{.98\columnwidth}{!}{
\begin{tabular}{c|cccc|cc} 
\hline
\multirow{2}{*}{Setting} & \multicolumn{4}{c|}{IoU}      & \multicolumn{2}{c}{Center}  \\ 
\cline{2-7}
                        & AR    & AP   & AP50  & AP75  & Recall & AP                 \\ 
\hline
ImageNet pretraining     & 0.693 & 0.601 & 0.899 & 0.676 & 0.963  & 0.922              \\
RICO-PW pretraining      & 0.714 & 0.634 & 0.916 & 0.741 & 0.961  & 0.938              \\
\hline
\end{tabular}

}
\caption{The detecting results of \emph{Graphic Pixel-Words} on P2S-UI with ImageNet or RICO-PW pretraining.
IoU and Center metrics are used for evaluation.
\label{tb:det_pretrain}}
\vspace{-5mm}
\end{table}

\begin{table}[t]
\centering
\begin{tabular}{c|c||c|c} 
\hline
Category        & top1 acc                  & Category             & top1 acc  \\ 
\hline
add             & 0.944                     & favorite             & 0.962     \\
arrow\_backward & 0.995                     & filter               & 0.880     \\
arrow\_downward & 0.981                     & gallery              & 0.930     \\
arrow\_forward  & 0.977                     & location             & 0.940     \\
arrow\_upward   & 0.876                     & menu                 & 0.985     \\
avatar          & 0.904                     & microphone           & 0.945     \\
calendar        & 0.897                     & more                 & 0.978     \\
call            & 0.930                     & other                & 0.929     \\
camera          & 0.900                     & pause                & 0.855     \\
cart            & 0.947                     & play                 & 0.974     \\
chat            & 0.937                     & question\_mark       & 0.968     \\
check           & 0.966                     & refresh              & 0.965     \\
close           & 0.951                     & search               & 0.941     \\
delete          & 0.896                     & send                 & 0.941     \\
download        & 0.998                     & settings             & 0.905     \\
edit            & 0.942                     & share                & 0.990     \\ 
\hline
Average         & \multicolumn{1}{c}{0.957} & \multicolumn{1}{c}{} &           \\
\hline
\end{tabular}

\caption{Top1 accuracy of our Graphic Classifier trained on RICO-ICON.\label{tb:graphic_classfication}}
\vspace{-7mm}
\end{table}



\section{Experiment}
\subsection{Experimental Setup}

\vspace{1mm}
\noindent
\textbf{Datasets}\;
\textbf{RICO\cite{deka2017rico, liu2018learning}} is a public dataset on mobile UI. There are totally 66,261 screenshots with corresponding metadata. The dataset is first proposed by Deka et al.\cite{deka2017rico} with screenshots and VH files only and Liu et al.\cite{liu2018learning} add semantic labels for every nodes of VH. 
\textbf{RICO-PW} is a dataset built by us for \emph{Pixel-Words} detection based on the public RICO dataset. (see section~\ref{sec:PW}) We take 67.5\%, 7.5\% and 25\% of screenshots for graphic detector's training, validation and testing respectively. During pretraining, we combine the training set and test set to obtain a better pretrained detector.
\textbf{P2S-UI} is a dataset collected by ourselves. This dataset contains 27077 mobile UI images, where 5556 images are annotated with \emph{Pixel-Words} manually. There also are the annotations for relation, clickablility and app type for each screenshot.
\textbf{RICO-ICON} is an icon dataset consisting of 120,067 icons cropped from the RICO dataset. The icons are divided into 32 categories.

\vspace{1mm}
\noindent
\textbf{Baselines}\;
\textbf{Leaf nodes} are used as the input of Transformer in the previous work~\cite{he2020actionbert}, which can be regarded as the baseline of \emph{Pixel-Words}.  
When extracting the leaf nodes, we use heuristic rules (e.g. spatial size and ratio of the nodes) to filter out some invalid leaf nodes. \textbf{W/o pretraining} is the baseline that the models trained without the RICO-PW dataset to study the impact of pretraining on the RICO-PW dataset. \textbf{W/o Screen Transformer} is the baseline to study the impact of our proposed Screen Transformer.


\begin{figure*}[t]
\centering
\includegraphics[width=0.7\textwidth]{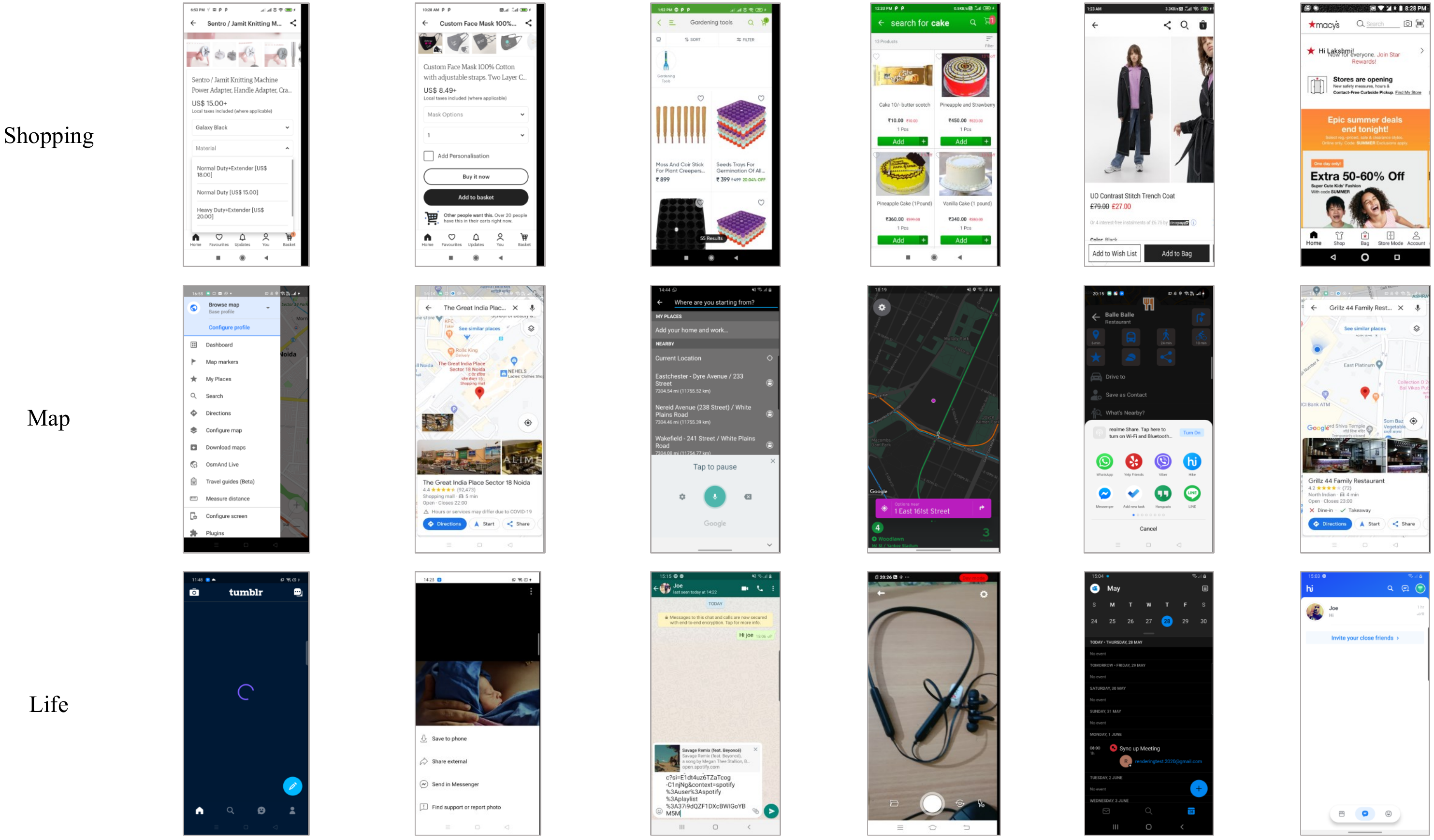} 
\caption{{ Visualization of the app type classification results on P2S-UI. The screenshots of the top/middle/bottom rows are classified as Shopping/Map/Life types respectively.}}
\label{fig:clsviz}
\vspace{-3mm}
\end{figure*}

\vspace{1mm}
\noindent
\textbf{Metrics}\;
To evaluate the performance of \emph{Pixel-Words} detection, we use Intersection over Union (IoU)-based detection metrics, which includes Average Recall (AR) of COCO-style~\cite{lin2014microsoft}, Average Precision (AP) of COCO-style, Average Precision with 0.50 IoU threshold (AP50), and Average Precision with 0.75 IoU threshold (AP75). We also use "Center" metric referring to \cite{zhang2021screen}. Given a predicted bounding box, the criteria of true or false is whether the center of the bounding box is inside the ground-truth box or not. For graphic classification, clickability prediction and relation prediction tasks, we use top1 accuracy as our metric.

\subsection{\emph{Pixel-Words} Evaluation}

In this section, we evaluate the extracting of \emph{Pixel-Words} from screenshots on RICO-PW and P2S-UI.

\vspace{1mm}
\noindent
\textbf{Study of Different Detection Models}\;
We select several typical one-stage detectors (for efficient inference) equipped with FPN to study the impact of different detectors on RICO-PW dataset. RetinaNet\cite{lin2017focal} is a generic one-stage detector, which is equipped with focal loss to solve the imbalance between positive and negative samples and assign labels on the basis of hand-crafted IoU threshold. ATSS\cite{zhang2020bridging} proposes a new mechanism which can adaptively adjust the IoU threshold to assign label. FreeAnchor\cite{zhang2021learning} treats the label assignment problem as a maximum estimation problem, which achieves the "learning to assign" during the training progress. Table~\ref{tb:det_baselines} shows results from above detectors. FreeAnchor obtains the highest AR and AP score, which means it performs well on both recognition and localization. One possible reason is that different from natural images, there are usually a lot of blank or texture-less backgrounds in the target bounding boxes in screenshots. FreeAnchor can better handle this problem and find the informative anchors to represent the detection target.
Therefore We use the FreeAnchor detector in following experiments.

\begin{figure*}[t]
\centering
\includegraphics[width=0.85\textwidth]{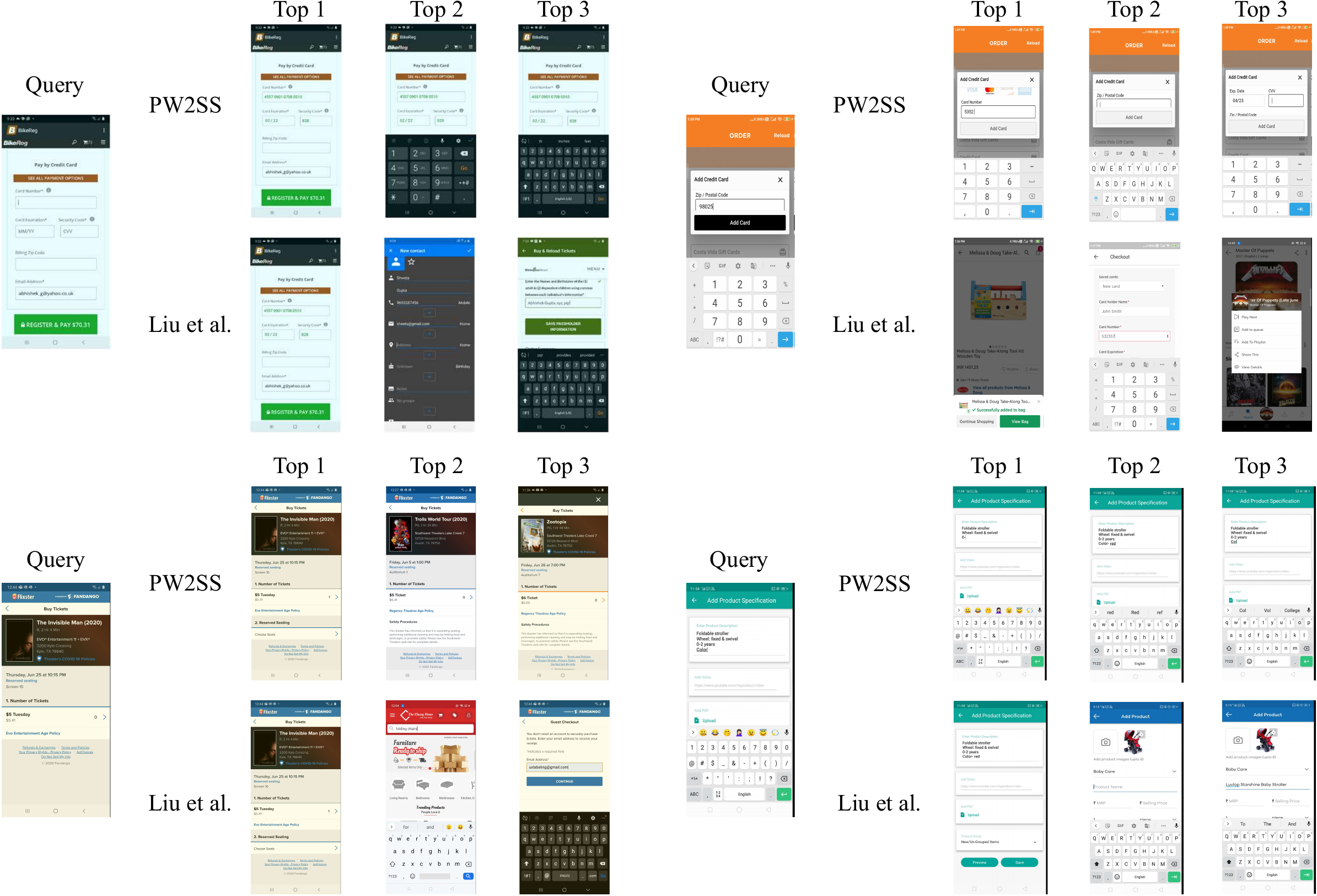} 
\caption{Comparison of the retrieval results of our PW2SS and Liu et al. ~\cite{liu2018learning} on P2S-UI. The retrieved screenshots of PW2SS are more similar to the query screenshots in terms of both semantics and layout.}
\label{fig:retieval}
\end{figure*}

\vspace{1mm}
\noindent
\textbf{\emph{Pixel-Words} v.s. Leaf Nodes}\;
We study which type of "word" of screenshots (\emph{Pixel-Words} or leaf nodes) is easier to be extracted. 
The \emph{Pixel-Words} and leaf nodes are learnt with the supervision of the cleaned leaf nodes and \emph{Pixel-Words} annotations respectively on RICO-PW. Visualized comparison are shown in Figure~\ref{fig:vis_det}.
Compared to leaf nodes, as Table~\ref{tab:leafnode_det} shows, our \emph{Pixel-Words} gain 17.29\%, 10.60\%, 15.87\%, 22.09\% on the metrics of AR, AP, AP50, AP75 respectively. The result supports our assumption that \emph{Pixel-Words} are easier to be extracted from pixels than leaf nodes and better fit our PW2SS framework.

\vspace{1mm}
\noindent
\textbf{Study of pretraining on RICO-PW}\;
We take our RICO-PW dataset as a pretraining dataset and finetune the detector on the human labeled P2S-UI dataset. Here we take the common ImageNet~\cite{imagenet_cvpr09} pretraining as baseline. Table~\ref{tb:det_pretrain} shows our results. Compared to ImageNet pretraning, our RICO-PW pretraining gains 3.03\%, 5.49\%, 1.89\%, 9.62\%  on AR, AP, AP50, AP75 respectively. The biggest improvement comes from AP75, which means the pretraining helps to localize graphics more precisely.

\vspace{1mm}
\noindent
\textbf{Study of Graphic Classifier}\;
The graphics play an important role in GUI, which can provide rich semantic meaning in limited pixels. We try to recognize the graphics as fine-grained categories to represent the semantic meanings. As mentioned in Section~\ref{sec:PW}, we reorganize the categories from original RICO dataset and train our graphic classifier on that. Here we use MobilenetV2 \cite{sandler2018mobilenetv2} as our Graphic Classifier. We report our category-level top1 accuracy in Table~\ref{tb:graphic_classfication}. The results show our Graphic Classifier can get rich information from our dataset. We achieve 0.975 average top1 accuracy of 32 categories.   

\subsection{PW2SS Evaluation}

We first pretrain the Screen Transformer with the masked \emph{Pixel-Words} prediction task, then we evaluate the effectiveness of PW2SS on \emph{Pixel-Word} level tasks (relation prediction and clickability prediction) and \emph{Screen-Sentence} level tasks (screen and app classification).

\vspace{1mm}
\noindent
\textbf{Pretraining on RICO-PW}\;
Before we train our Screen Transformer in downstream tasks, we first pretrain it on the RICO-PW dataset. The goal of pretraining is to force the Screen Transformer to model relations between \emph{Pixel-Words} using the task of Mask \emph{Pixel-Words} Model. For each screenshot, 15\% of \emph{Pixel-Words} are randomly masked. We pretrain the model for 50 epochs, and using 5 epoch for warming up. We use the AdamW optimizer with learning rate $r=10^{-4}$, $\beta_1=0.9$,$\beta_2=0.999$, $\epsilon=10^{-6}$ and batch size of 64 for training. 

\vspace{1mm}
\noindent
\textbf{Clickability Prediction}\;
In this task, we add a classifier head on the output feature of the Screen Transformer to predict whether a \emph{Pixel-Word} is clickable or not. Table \ref{tb:rpcp} shows that the with the Screen Transformer, we can achieve better clickability results using the context of \emph{Pixel-Words}. Besides, the pretraining on RICO-PW helps to further improve the accuracy, which indicates that the masked \emph{Pixel-Words} task is effective to learn a better representation by modeling the relation between \emph{Pixel-Words}.


\vspace{1mm}
\noindent
\textbf{Relation Prediction}\;
Relation prediction task is designed to evaluate the performance of our Screen Transformer on the tasks about \emph{Pixel-Words} pairs. Examples of the relation are shown in Figure~\ref{fig:RP}. As Table \ref{tb:rpcp} shows, similar to the clickability prediction task, both the Screen Transformer and pretraining are effective for these tasks. 
An interesting observation is that, compared to the clickability prediction task, the gain caused by Screen Transformer in relation prediction task is larger. The reason is that relation prediction task relies more on the context of \emph{Pixel-Words}, where the Screen Transformer can provide more useful information.

\begin{table}[t]
\centering
\resizebox{.95\columnwidth}{!}{
\begin{tabular}{c|c|c} 
\hline
Setting                                & CP    & RP     \\ 
\hline
w/o Screen Transformer                  & 0.871 & 0.906  \\
w/ Screen Transformer (w/o pretraining) & 0.896 & 0.947  \\
w/ Screen Transformer (w/ pretraining)  & \textbf{0.910} & \textbf{0.965}  \\
\hline
\end{tabular}}
\caption{Results of Clickability Prediction (CP) and Relation Prediction (RP) on P2S-UI. We study the impact of using Screen Transformer and pretraining on RICO-PW.}
\label{tb:rpcp}
\vspace{-7mm}
\end{table}

\vspace{1mm}
\noindent
\textbf{App Type Classification}\;
All the screenshots in P2S-UI dataset are divided into the 26 categories according to the app type (e.g. SHOP, SOCIAL). We learn to predict the app type for each \emph{Screen-Sentence}. As Table \ref{tb:app} shows, our \emph{Pixel-Words} outperform leaf nodes with more than 3\% gain, which verifies the effectiveness of our defined \emph{Pixel-Words} in screen-level understanding. The qualitative results are shown in Figure ~\ref{fig:clsviz}.


\begin{table}[t]
\centering
\resizebox{.85\columnwidth}{!}{
\begin{tabular}{c|c|c} 
\hline
Setting                              & model                  & accuracy  \\ 
\hline
\multirow{2}{*}{Leaf Node}            & w/o Screen Transformer & 0.914   \\
                                      & w/ Screen Transformer  & 0.915  \\ 
\hline
\multirow{2}{*}{\textit{Pixel-Words}} & w/o Screen Transformer & 0.948   \\
                                      & w/ Screen Transformer  & 0.955 \\
\hline
\end{tabular}}
\caption{App type classification results on P2S-UI. We study the impact of different inputs of Screen Transformer: \emph{Pixel-Words} v.s. Leaf Nodes, and the impact of Screen Transformer.}
\vspace{-7mm}
\label{tb:app}
\end{table}


\vspace{1mm}
\noindent
\textbf{Screen Retrieval}\;
Screen Retrieval is a well-applied task to evaluate the models' ability to learn the representation of the whole screen. Given the query screen, we will retrieval We compare our model to the method\cite{liu2018learning}, which uses a autoencoder of the screen layout to retrieval screens. As Figure~\ref{fig:retieval} shows, the screen embedding generated by Screen Transformer is more {suitable} to retrieve the screens having similar semantic meaning as well as spatial layout. 

\section{Conclusion}
In this work, we propose a pixel-based GUI understanding framework, PW2SS, which is suitable for general applications across different platforms, UI design tools and styles. We extend the \emph{Word}/\emph{Sentence} concepts into the \emph{Pixel-Word}/\emph{Screen-Sentence} concepts in the GUI understanding area. \emph{Pixel-Words} are defined as atomic components with essential and clear semantics. Based on the definition of the \emph{Pixel-Words}, we can use OCR and  graphic detector to extract the \emph{Pixel-Words} from the screenshots. Then a Screen Transformer is proposed to model the relation between \emph{Pixel-Words}.  The effectiveness of PW2SS is verified in tasks including \emph{Pixel-Words} extraction, relation prediction, clickability prediction, screen retrieval, and app type classification.
We can also extend our work to understanding screenshot sequences of user accomplishing various tasks by aggregating \emph{Screen-Sentences} into \emph{Task-Paragraphs}.
\bibliographystyle{ACM-Reference-Format}
\bibliography{main}











\end{document}


\title{Supplementary Material of
Understanding Mobile GUI: from Pixel-Words to Screen-Sentences}






\author{Anonymous authors}
\affiliation{
 \institution{Paper under double-blind review}
 \country{}
}
\renewcommand{\shortauthors}{Anonymous Author, et al.}

\maketitle

\begin{figure}[t]
\centering
\includegraphics[width=0.48\textwidth]{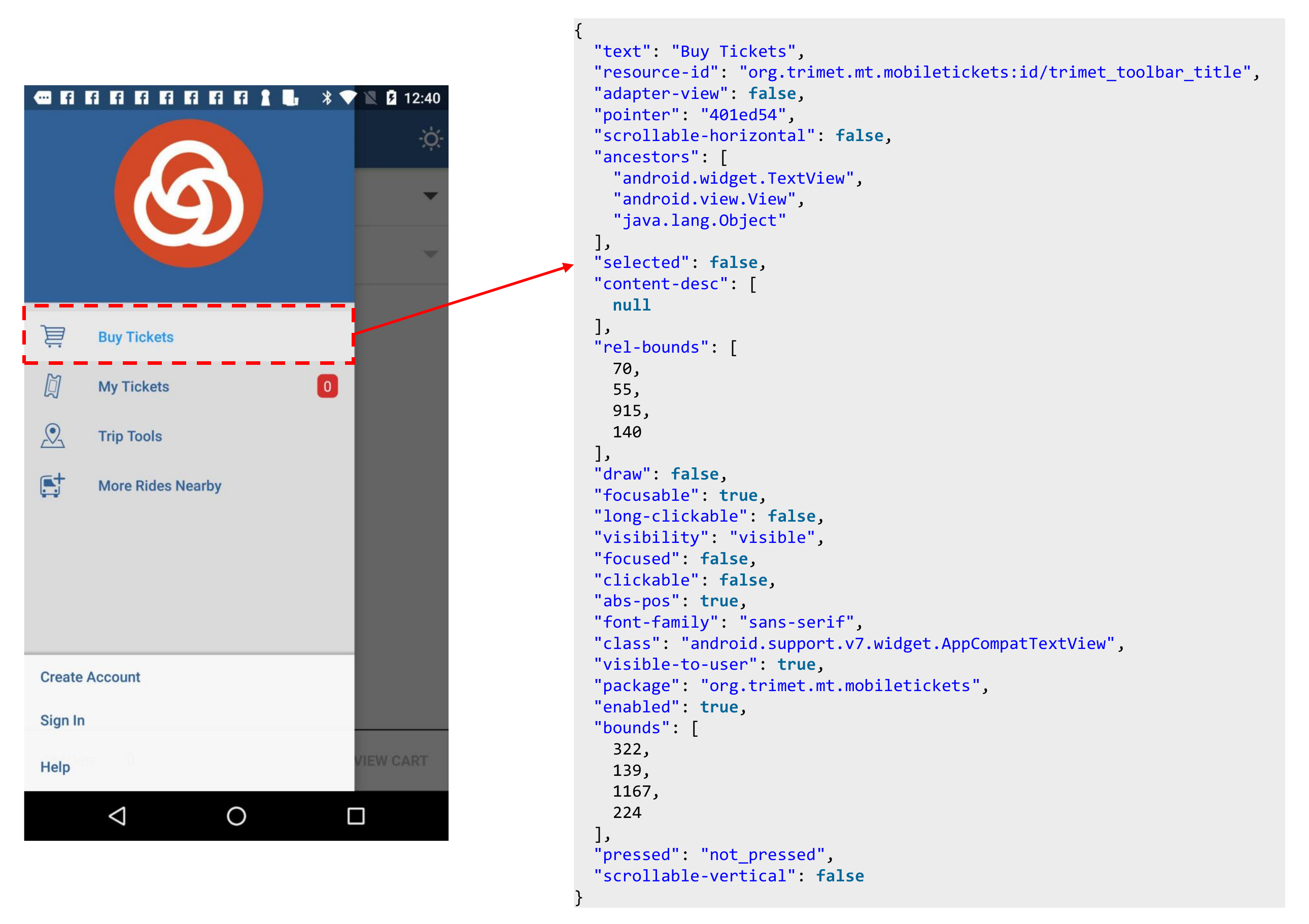}
\caption{ A typical node data in View Hierarchy (VH) from RICO dataset.
}
\label{fig:example_vh}
\end{figure}

\section*{Details of \emph{Pixel-Words} annotation generation}

\begin{figure}[t]
\centering
\includegraphics[width=0.48\textwidth]{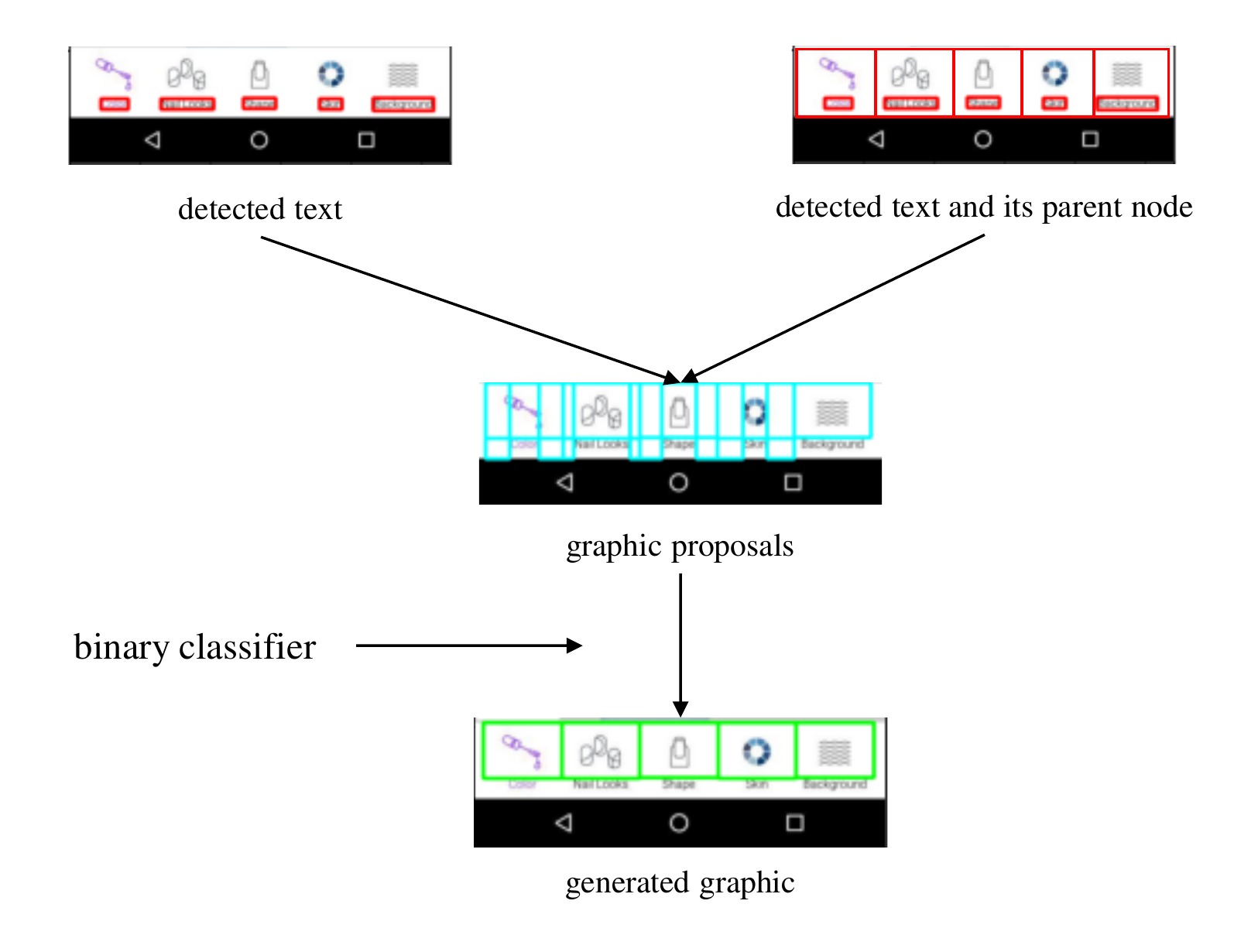} 
\caption{ The visualization of progress to generate missing graphic in View Hierarchy (VH).
}
\label{fig:gen_graphic}
\end{figure}

To solve the invalid node problem in VH, motivated by “proposal-classification”\cite{girshick2014rich, girshick2015fast, shaoqing2015faster} style detection method. We try to extract proposals from the VH then use a classifier to identify each proposal is a graphic or not.

To train such a binary classifier, we collect 1273 patches of VH nodes from the original screenshots of RICO. 694 of them are labeled as positive samples and the rest 539 patches are labeled as negative samples. After training, our classifier achieves 0.95 accuracy.

However, there are two challenges here. First, there are a huge number of nodes in VH, if we take every node as a proposal it will be pretty inefficient and increase the number of failure cases for our classification. Second, about 28\% of the graphics don’t have corresponding nodes in VH, so it means the upper bound of recall will be very limited. To tackle the first challenge, we create a candidate set of node class names that covers all graphics-related categories. When we sample nodes from VH, only the node whose class name can be find in the candidate set will be considered as possible proposals for graphics. For the second problem, we observed that the missing graphic nodes often associate with texts as Figure ~\ref{fig:example_vh} shows. Therefore we generate new proposals according to the texts and their parent nodes. Specifically, for a text and its parent node, we select the regions between the text and its parent node’s top/bottom/left/right boundaries as proposals respectively, as shown in Figure~\ref{fig:gen_graphic}. The methods solving the first and second challenge helps to improve the precision and recall respectively.

\begin{lstlisting}[language=Python, caption=pseudo code for \emph{Graphic Pixel-Words} Annotation Generation]
# nodes: List[Dict], all nodes data in a VH file 
# clsname_candidates: Set(str), a set which includes possible graphic class in VH.
# text_bboxes: List[List[int]], the list of text's bounding boxes in current screenshots.

proposals = []
# step1: filter out unrelated nodes.
for node in nodes:
    if node["class"] in clsname_candidates or node["ancestors"] in clsname_candidates:
        proposals.append(node["bound"])    

# step2: add spaced_region between text and its parent node.
for text in text_bboxes:
    parent = get_parent_node_bbox(text)
    proposals += generate_spaced_region(parent, text)

# step3: use the binary classifier to evaluate the proposals.
graphics_bboxes = []
for proposal in proposals:
    if evaluate(proposal) > score_thres:
        graphics_bboxes.append(proposal)
return graphics_bboxes
\end{lstlisting}


\bibliographystyle{ACM-Reference-Format}
\bibliography{main}